\documentclass{article}
\usepackage{spconf,amsmath,graphicx,hyperref,pifont}
\usepackage[utf8]{inputenc}
\usepackage{booktabs} 
\usepackage{multirow} 
\usepackage{graphicx} 
\usepackage{caption}
\usepackage{float}
\usepackage{graphicx} 
\usepackage{subcaption} 
\usepackage{caption}
\usepackage{amssymb}  
\usepackage[table]{xcolor}
\usepackage{comment} 
\usepackage{algorithm}
\usepackage{algorithmicx}
\usepackage{algpseudocode}
\usepackage{setspace}

\title{A Two-Stage Globally-Diverse Adversarial Attack for Vision-Language Pre-training Models
}
\name{Wutao Chen$^{1}$, Huaqin Zou$^{1}$, Chen Wan$^{1}$\textsuperscript{*}, Lifeng Huang$^{2}$ 
\thanks{\textsuperscript{*}Corresponding author.}
}
\address{$^{1}$Department of Computer Science and Technology, Shantou University, Shantou, China\\
$^{2}$College of Mathematics and Informatics, South China Agricultural University, Guangzhou, China}
\DeclareMathOperator*{\argmax}{arg\,max}
\begin{document}

%
\maketitle
  \begin{abstract}
Vision-language pre-training (VLP) models are vulnerable to adversarial examples, particularly in black-box scenarios. Existing multimodal attacks often suffer from limited perturbation diversity and unstable multi-stage pipelines. To address these challenges, we propose 2S-GDA, a two-stage globally-diverse attack framework. The proposed method first introduces textual perturbations through a globally-diverse strategy by combining candidate text expansion with globally-aware replacement. To enhance visual diversity, image-level perturbations are generated using multi-scale resizing and block-shuffle rotation. Extensive experiments on VLP models demonstrate that 2S-GDA consistently improves attack success rates over state-of-the-art methods, with gains of up to 11.17\% in black-box settings. Our framework is modular and can be easily combined with existing methods to further enhance adversarial transferability.
\end{abstract}
\section{Introduction}
\label{sec:intro}

Vision-language pre-training (VLP) models have achieved impressive performance across a wide range of multimodal tasks, such as image-text retrieval~\cite{khan2021exploiting} and visual question answering~\cite{yang2022vision}. Despite their success, recent studies~\cite{zhang2022towards,lu2023set} have revealed that these models are vulnerable to adversarial examples. By introducing small perturbations to either the image, the text, or both modalities, adversarial examples can disrupt multimodal semantic alignment and lead to substantial performance degradation~\cite{xu2024image,zhou2024improving}. Moreover, adversarial examples generated by a surrogate model can remain effective in black-box settings, revealing their cross-model transferability~\cite{11045302,gao2024boosting}. This property poses a practical threat in black-box scenarios and has attracted growing attention~\cite{luo2024image}.

To enhance adversarial transferability in multimodal settings, recent works have explored joint perturbation strategies over image-text pairs. Co-Attack~\cite{zhang2022towards} performs coordinated optimization on single pairs, achieving strong white-box performance but limited transferability due to insufficient diversity. Set-level Guidance Attack (SGA)~\cite{lu2023set} addresses this issue by introducing a three-stage framework with set-level textual optimization and multi-scale image augmentations. Wang et al.~\cite{KYXH202502012} incorporate BSR (block shuffling and rotation) augmentation strategy~\cite{bsr} into SGA to enhance visual perturbations. DRA~\cite{gao2024boosting} enriches adversarial trajectories through multi-step perturbation modeling, and SA-AET~\cite{11045302} leverages adaptive augmentations to improve transferability.

\begin{figure}[!]
\centering
\subcaptionbox{SGA\label{fig:1a}}
{\includegraphics[width=.22\textwidth]{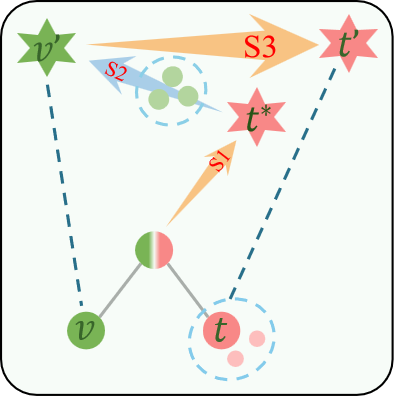}}
\hspace{1em}
\subcaptionbox{2S-GDA\label{fig:1b}}
{\includegraphics[width=.166\textwidth]{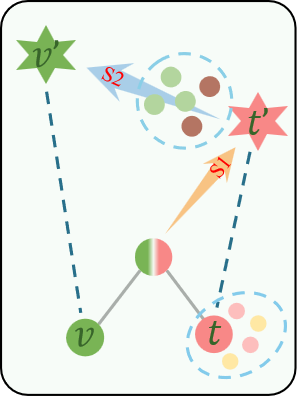}}
\vspace{-1mm}
\caption{
Comparison of cross-modal interactions. (a) \textbf{SGA}: three-stage perturbation (S1: text~$\rightarrow$~S2: image~$\rightarrow$~S3: text). (b) \textbf{2S-GDA}: two-stage perturbation (S1: text $\rightarrow$ S2: image). $v$: input image; $t$: paired caption; $t^{*}$: intermediate state; $v'$, $t'$: corresponding adversarial examples. Arrows indicate guidance for generating adversarial examples.}
\label{fig:1}
\vspace{-5mm}
\end{figure} 

\begin{figure}[t]
\centering
\subfloat[ASR comparison across stages]{\includegraphics[width=.48\textwidth]{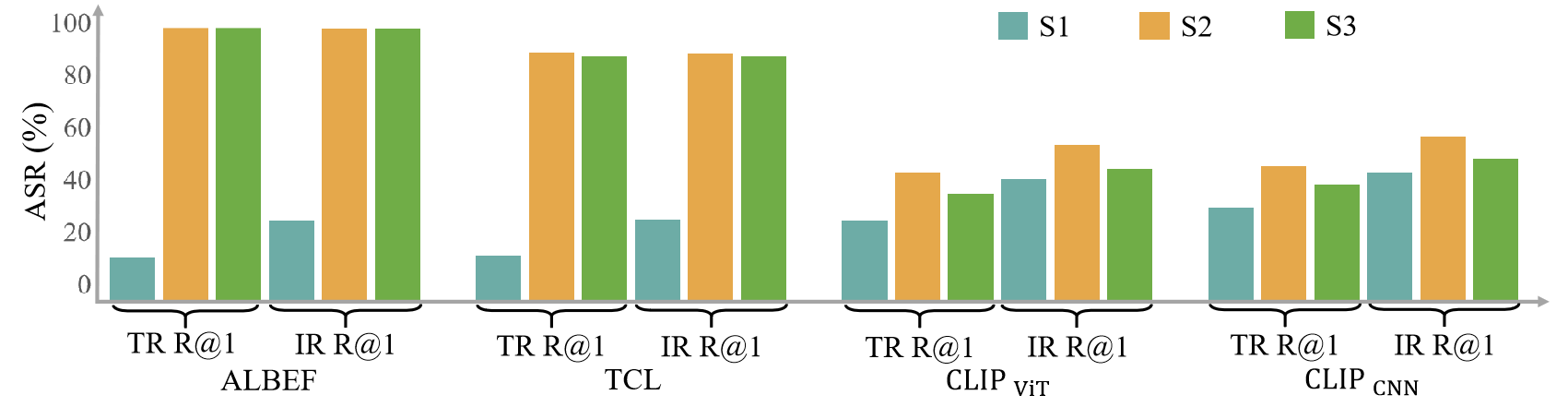}
\label{fig:2a}}
\vspace{-0.3mm}
\subfloat[Euclidean distance]{
  \includegraphics[width=0.22\textwidth]{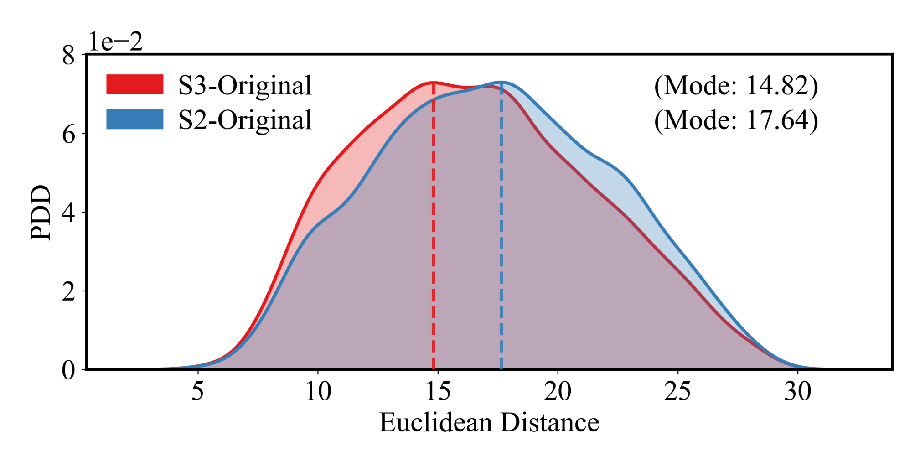}
  \label{fig:2b}
}
\hfill
\subfloat[Cosine similarity]{
  \includegraphics[width=0.22\textwidth]{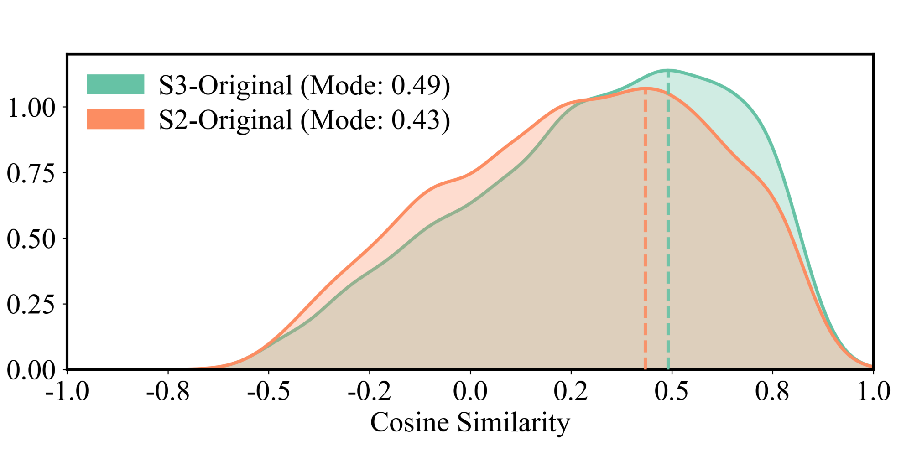}
  \label{fig:2c}
}
\vspace{-1mm}
\caption{Feature-space analysis of SGA.}
\label{fig:2}
\vspace{-5mm}
\end{figure}

However, two key limitations remain that may hinder the effectiveness of existing approaches. First, as illustrated in Fig.~\ref{fig:1a}, the representative framework SGA~\cite{lu2023set} adopts a three-stage pipeline (S1: text~$\rightarrow$~S2: image~$\rightarrow$~S3: text), where an initial discrete textual perturbation is followed by continuous image perturbation and a final round of text modification. The final stage often introduces semantic drift, which can undermine the effectiveness of earlier gains. As shown in Fig.~\ref{fig:2a}, the attack success rate (ASR) of S2 generally exceeds that of S3, demonstrating that the last-stage modification degrades overall effectiveness. We compute the probability density distributions (PDD) of Euclidean distance and cosine similarity in feature space between the original input and the adversarial examples generated by S2 and S3. As shown in Figs.~\ref{fig:2b} and~\ref{fig:2c}, adversarial examples from S3 are closer to the original input in the feature space, with a lower Euclidean distance (14.82 vs. 17.64) and higher cosine similarity (0.49 vs. 0.43), indicating reduced feature-level deviation and weaker perturbation (as detailed in the supplementary material). Second, most existing methods rely on greedy substitution strategies that select only the top-1 token from BERT-MLM~\cite{devlin2019bert} predictions. This heuristic overlooks the semantic contributions of non-key tokens and constrains the exploration of a more globally diverse perturbation space.

To address these challenges, we propose \textbf{2S-GDA (Two-Stage Globally-Diverse Attack)}, a new attack framework that improves transferability by rethinking both the attack pipeline structure and the semantic perturbation mechanism. As illustrated in Fig.~\ref{fig:1b}, 2S-GDA restructures the process into two stages, avoiding the unstable re-perturbation of the text. Specifically, we introduce a globally-aware textual perturbation module that constructs a multi-source candidate pool using BERT-MLM and WordNet~\cite{miller1995wordnet}, and selects replacements by jointly considering token importance and cross-modal alignment loss. In parallel, a visual perturbation module applies multi-scale resizing and BSR~\cite{bsr} to enhance the diversity and effectiveness of image-level adversarial perturbations.
Extensive experiments on VLP models demonstrate that 2S-GDA consistently outperforms state-of-the-art baselines, particularly in black-box settings. Moreover, the proposed framework is modular and extensible, and can be easily integrated into existing methods to further enhance adversarial transferability.
We summarize our contributions as follows.
\begin{itemize}
    \item We propose 2S-GDA, a new two-stage attack framework that eliminates the instability introduced by repeated textual perturbations.
    \item We design a globally-aware textual perturbation mechanism that expands candidate space and avoids local greedy traps.
    \item Extensive experiments demonstrate that our method consistently outperforms state-of-the-art baselines, especially under black-box settings.
\end{itemize}

\section{Methodology}
\subsection{Notations}
Let $(v,t)$ denote an image-text pair sampled from a multimodal dataset $\mathcal D$. In VLP models, we denote the image encoder as $\mathcal{F}_I(\cdot)$, the text encoder as $\mathcal{F}_T(\cdot)$, and the multimodal fusion module as $\mathcal{F}_M(\cdot)$. Specifically, the image and text embeddings are obtained as $e_v = \mathcal{F}_I(v)$ and $e_t = \mathcal{F}_T(t)$, respectively. The fused multimodal representation is then computed as $\mathcal{F}_M(e_v, e_t)$.

We define $\mathcal{B}[v, \varepsilon_v]$ and $\mathcal{B}[t, \varepsilon_t]$ as the adversarial search spaces for the image and text modalities, respectively. Here, $\varepsilon_v$ denotes the maximum allowable perturbation magnitude (e.g., $\ell_\infty$ norm bound), and $\varepsilon_t$ denotes the maximum number of token substitutions permitted within the caption. 
The goal is to craft adversarial examples $(v', t') \in \mathcal{B}[v, \varepsilon_v] \times \mathcal{B}[t, \varepsilon_t]$ such that the fused multimodal embedding $\mathcal{F}_M(\mathcal{F}_I(v'), \mathcal{F}_T(t'))$ becomes misaligned with the clean representation, resulting in incorrect retrieval results.

\subsection{Two-Stage Attack Pipeline}
To overcome the limitations of existing three-stage frameworks (e.g., SGA), 2S-GDA introduces a streamlined two-stage pipeline that removes the unstable textual re-perturbation phase while retaining both semantic guidance and perturbation diversity. The proposed framework consists of the following two core stages.


\noindent \textit{\textbf{Stage I: Textual Perturbation}}

Given a caption $t$, we utilize its paired image encoding $\mathcal{F}_I({v})$ as cross-modal supervision to identify a subset of critical words $t' = \{t'_{1}, t'_{2}, \dots, t'_{M}\}$ that are most crucial for cross-modal matching. For each selected token, a set of candidate replacements is generated using a Masked Language Model (MLM). The objective is to reduce the semantic consistency between the perturbed text and the original image by minimizing the cosine similarity between the perturbed text embedding $\mathcal{F}_T(t')$ and the original image features $\mathcal{F}_I(v)$:
\begin{equation} \label{eq:1}
t' = \argmax_{t' \in \mathcal{B}[t, \varepsilon_t]} 
\left( -\frac{\mathcal{F}_T(t') \cdot \mathcal{F}_I(v)}
{\|\mathcal{F}_T(t')\| \|\mathcal{F}_I(v)\|} \right)
\end{equation}
\noindent  where $\mathcal{F}_T$ and $\mathcal{F}_I$ denote the text and image encoders, respectively. The adversarial caption $t'$ is optimized to be maximally distant from the original image $v$ in the embedding space.

\noindent \textit{\textbf{Stage II: Visual Perturbation}}

In this stage, the goal is to introduce adversarial perturbations to the image $v$ such that the resulting visual embedding $\mathcal{F}_I(v')$ is misaligned with the adversarial text $t'$ obtained in Stage I.
To increase attack generalization, we apply a set of transformation-based augmentations, including multi-scale resizing and BSR, to construct a diverse image set $\mathcal S_v$ for adversarial optimization. The adversarial image $v'$ is generated by solving the following optimization problem:
\begin{equation} \label{eq:e}
v' = \underset{v' \in B[v,\varepsilon_v]}{\arg\!\max} 
-\sum_{i=1}^{M} 
\frac{\mathcal{F}_T(t'_i)}{\|\mathcal{F}_T(t'_i)\|} 
\sum_{v_i \in S_v} 
\frac{\mathcal{F}_I(v_i)}{\|\mathcal{F}_I(v_i)\|}
\end{equation}

\noindent where $S_v$ denotes the set of image variants generated by applying BSR and multi-scale resizing to $v'$, and each $v_i \in S_v$ is obtained by scaling $v'$ with a factor $s_i$.
This objective encourages all $v_i \in S_v$ to remain semantically distant from adversarial captions $t'_i$ in the embedding space.


\subsection{Globally-Diverse Strategy}
To increase semantic diversity and avoid overfitting to the top-1 token, a Globally-Diverse Strategy is introduced, consisting of two components.

\noindent \textit{1) Candidate Text Expansion (CTE)}.
Since standard MLMs provide limited variations, WordNet is incorporated to build a multi-source candidate pool. For each token $w$ selected for perturbation, the top-$k$ predictions from BERT-MLM ($C_{\text{MLM}}(w)$) are combined with synonyms from WordNet ($C_{\text{WN}}(w)$). The final candidate set excludes invalid items such as stopwords, subword fragments, and special symbols,
\begin{equation} \label{eq:4}
C(w)=\{c \in C_{\text{MLM}}(w) \cup C_{\text{WN}}(w) | c\neq  R\}
\end{equation}
\noindent where $R$ denotes the set of excluded items. 
This expansion enlarges the perturbation space while preserving semantic plausibility.


\noindent \textit{2) Globally-Aware Replacement (GAR)}.
We employ a global scoring mechanism to evaluate the impact of candidate substitutions on cross-modal alignment. Specifically, the top-$k$ most influential tokens in the input text $t$ are identified by gradient-based saliency scoring on a masked language model (MLM), forming the set ${W}_{\text{imp}} = \{w_1, w_2, \dots, w_k\}$, where $k$ is the number of influential tokens.
For each $w_i \in W_{\text{imp}}$, the candidate set $C(w_i)$ is constructed according to Eq.~(\ref{eq:4}). Each candidate token $\tilde{w} \in C(w_i)$ is then substituted into $t$, and the corresponding adversarial loss $\mathcal{J}(v, t : w_i \rightarrow \tilde{w})$ is computed. We select the token–candidate pair that yields the maximum adversarial loss,
\begin{equation} \label{eq:optimal_t}
(w_i' \rightarrow \tilde{w}') = 
\mathop{\mathrm{arg\,max}}_{\substack{w_i \in W_{\text{imp}} \\ \tilde{w} \in C(w_i)}}
\mathcal{J}(v, t : w_i \rightarrow \tilde{w})
\end{equation}
where $W_{\text{imp}}$ denotes the set of important tokens and $C(w_i)$ represents the candidate pool for $w_i$.
Note that although $k$ influential tokens are initially selected, only $\varepsilon_t$ token with the highest alignment disruption is ultimately replaced. This selective strategy preserves semantic stability and mitigates excessive semantic drift.

The detailed 2S-GDA algorithm is provided in the supplementary material. Compared to existing methods (e.g., SGA and its variants), our approach removes the unstable text re-perturbation stage and introduces globally-diverse textual augmentation, resulting in more semantically consistent and transferable adversarial examples.

\setlength{\tabcolsep}{2pt}
\begin{table}[t]
    \centering
  
    \caption{
        ASR (\%) of our method and the baseline approaches. Diagonal entries indicate white-box attacks.}
        
    \fontsize{5.5pt}{7.5pt}\selectfont 
    \begin{tabular}{cccccccccc}
        
        \toprule
      
        \multirow{2}{*}{Source} & \multirow{2}{*}{Attack} & \multicolumn{2}{c}{{ALBEF}} & \multicolumn{2}{c}{{TCL}} & \multicolumn{2}{c}{{CLIP}\textsubscript{ViT}} & \multicolumn{2}{c}{{CLIP}\textsubscript{CNN}} \\
        \cmidrule(lr){3-4} \cmidrule(lr){5-6} \cmidrule(lr){7-8} \cmidrule(lr){9-10}
    
        & & TR R@1 & IR R@1 & TR R@1 & IR R@1 & TR R@1 & IR R@1 & TR R@1 & IR R@1 \\
        \midrule

        \multirow{5}{*}{{ALBEF}} 
        & SGA~\cite{lu2023set}     & \cellcolor{gray!10}\textbf{100.00} & \cellcolor{gray!10}99.95 & 89.67 & 89.64 & 39.14 & 47.87 & 42.66 & 52.14 \\
        & SGA-BSR~\cite{KYXH202502012} & \cellcolor{gray!10}\textbf{100.00} & \cellcolor{gray!10}99.95 & 94.20 & 94.33 & 63.68 & 70.01 & 71.26 & 73.34 \\
        & DRA~\cite{gao2024boosting}     & \cellcolor{gray!10}99.90 & \cellcolor{gray!10}99.93 & 91.68 & 91.86 & 47.12 & 57.18 & 51.72 & 60.07 \\
        & SA-AET~\cite{11045302} & \cellcolor{gray!10}99.90 & \cellcolor{gray!10}99.98 & 96.42 & 96.02 & 55.58 & 63.89 & 57.22 & 65.59 \\
        & 2S-GDA (ours) & \cellcolor{gray!10}\textbf{100.00} & \cellcolor{gray!10}\textbf{100.00} & \textbf{97.26} & \textbf{96.40} & \textbf{74.85} & \textbf{78.77} & \textbf{77.01} & \textbf{81.34} \\
        \midrule

        \multirow{5}{*}{{TCL}} 
        & SGA~\cite{lu2023set}      & 92.60 & 92.96 & \cellcolor{gray!10}\textbf{100.00} & \cellcolor{gray!10}\textbf{100.00} & 37.91 & 48.16 & 42.91 & 52.80 \\
        & SGA-BSR~\cite{KYXH202502012}  & 96.66 & 96.66 & \cellcolor{gray!10}\textbf{100.00} & \cellcolor{gray!10}99.95 & 64.66 & 69.33 & 69.99 & 74.58 \\
        & DRA~\cite{gao2024boosting}      & 94.68 & 95.28 & \cellcolor{gray!10}\textbf{100.00} & \cellcolor{gray!10}99.95 & 47.61 & 57.54 & 51.85 & 62.06 \\
        & SA-AET~\cite{11045302}      & \textbf{98.85} & 98.50 & \cellcolor{gray!10}\textbf{100.00} & \cellcolor{gray!10}\textbf{100.00} & 56.20 & 63.47 & 59.77 & 67.86 \\
        & 2S-GDA (ours) & 97.50 & \textbf{98.13} & \cellcolor{gray!10}\textbf{100.00} & \cellcolor{gray!10}\textbf{100.00} & \textbf{73.37} & \textbf{79.48} & \textbf{78.67} & \textbf{83.70} \\
        \midrule

        \multirow{5}{*}{{CLIP}\textsubscript{ViT}} 
        & SGA~\cite{lu2023set}      & 22.73 & 35.69 & 25.29 & 37.19 & \cellcolor{gray!10}\textbf{100.00} & \cellcolor{gray!10}\textbf{100.00} & 55.56 & 62.06 \\
        & SGA-BSR~\cite{KYXH202502012}  & 49.64 & 59.19 & 51.63 & 58.38 & \cellcolor{gray!10}\textbf{100.00} & \cellcolor{gray!10}\textbf{100.00} & 83.52 & 84.67 \\
        & DRA~\cite{gao2024boosting}      & 29.20 & 44.34 & 31.09 & 45.19 & \cellcolor{gray!10}\textbf{100.00} & \cellcolor{gray!10}\textbf{100.00} & 65.64 & 70.33 \\
        & SA-AET~\cite{11045302}      & 36.60 & 50.44 & 39.20 & 51.10 & \cellcolor{gray!10}\textbf{100.00} & \cellcolor{gray!10}\textbf{100.00} & 71.01 & 74.10 \\
        & 2S-GDA (ours) & \textbf{54.54} & \textbf{65.18} & \textbf{55.43} & \textbf{65.45} & \cellcolor{gray!10}\textbf{100.00} & \cellcolor{gray!10}99.97 & \textbf{90.04} & \textbf{89.98} \\
        \midrule

        \multirow{5}{*}{{CLIP}\textsubscript{CNN}} 
        & SGA~\cite{lu2023set}      & 15.54 & 29.58 & 18.12 & 32.57 & 42.58 & 52.71 & \cellcolor{gray!10}99.87 & \cellcolor{gray!10}\textbf{99.97} \\
        & SGA-BSR~\cite{KYXH202502012}  & 24.19 & 36.97 & 26.13 & 40.86 & 52.15 & 61.08 & \cellcolor{gray!10}\textbf{100.00} & \cellcolor{gray!10}99.93 \\
        & DRA~\cite{gao2024boosting}      & 19.50 & 34.47 & 21.18 & 37.71 & 48.34 & 59.79 & \cellcolor{gray!10}\textbf{100.00} & \cellcolor{gray!10}\textbf{99.97} \\
        & SA-AET~\cite{11045302}      & 23.98 & 38.28 & 27.29 & 41.81 & 54.11 & 64.21 & \cellcolor{gray!10}\textbf{100.00} & \cellcolor{gray!10}99.97 \\
        & 2S-GDA (ours) & \textbf{28.78} & \textbf{46.02} & \textbf{34.35} & \textbf{49.05} & \textbf{67.61} & \textbf{75.19} & \cellcolor{gray!10}\textbf{100.00} & \cellcolor{gray!10}99.90 \\
        \bottomrule
    \end{tabular}

    \label{tab:1}
    \vspace{-1mm}
\end{table}

\section{Experiments}
\subsection{Experimental Settings}
\noindent {\bf Datasets}. Experiments are conducted on the Flickr30K dataset~\cite{plummer2015flickr30k}, which contains 1000 images with five human-annotated captions each. Additional results on the MSCOCO dataset are provided in the supplementary material\footnote{The code and supplementary material are available at \url{https://github.com/chenwutao123/2S-GDA}.}.

\noindent {\bf Models}. Two types of VLP model are selected, i.e., fused and aligned. ALBEF~\cite{li2021align} and TCL~\cite{yang2022vision} represent fused models, while CLIP\textsubscript{ViT} and CLIP\textsubscript{CNN}~\cite{radford2021learning} serve as aligned models.

\noindent {\bf Baselines}. We compare our approach with four recent attack methods, i.e., SGA~\cite{lu2023set}, SGA-BSR~\cite{KYXH202502012}, DRA~\cite{gao2024boosting}, and SA-AET~\cite{11045302}, as well as their combined variants (DRA-BSR and SA-AET-BSR). Earlier methods such as PGD~\cite{madry2017towards}, BERT-Attack~\cite{li2020bert}, Sep-Attack, and Co-Attack~\cite{zhang2022towards} are reported in the supplementary material.

\noindent {\bf Attack Settings}. Following~\cite{11045302,gao2024boosting}, image perturbations are generated using PGD with $\varepsilon_v = 8/255$, step size $2/255$, and $T = 10$ iterations. For the text modality, we adopt BERT-Attack~\cite{li2020bert} with a perturbation bound $\varepsilon_t = 1$, a candidate list size of $W = 10$, and $k = 3$ influential tokens.

\noindent {\bf Metrics}. The effectiveness of adversarial attacks is evaluated using the Attack Success Rate (ASR) measured on Rank-1 retrieval accuracy, including both text-to-image (TR R@1) and image-to-text (IR R@1) tasks.

\vspace{-2mm}
\subsection{Experimental Results}
Table~\ref{tab:1} reports the ASR (\%) of our proposed 2S-GDA and four baselines (SGA, SGA-BSR, DRA, and SA-AET). Diagonal entries indicate white-box attacks, and off-diagonal entries reflect transferability. Compared to baselines, 2S-GDA generally achieves 100\% ASR in white-box settings, and consistently outperforms them in black-box settings, with improvements of 0.11\% $\sim$ 11.17\%. These results demonstrate the effectiveness of our method in enhancing adversarial transferability.

To evaluate the extensibility of our method, we integrate 2S-GDA into two baseline frameworks (DRA and SA-AET). Table~\ref{tab:2} demonstrates that our method consistently achieves 1.57\% to 7.51\% higher ASR than the baselines in black-box settings.

\begin{figure}[!]
    \centering
    \includegraphics[width=0.98\linewidth]{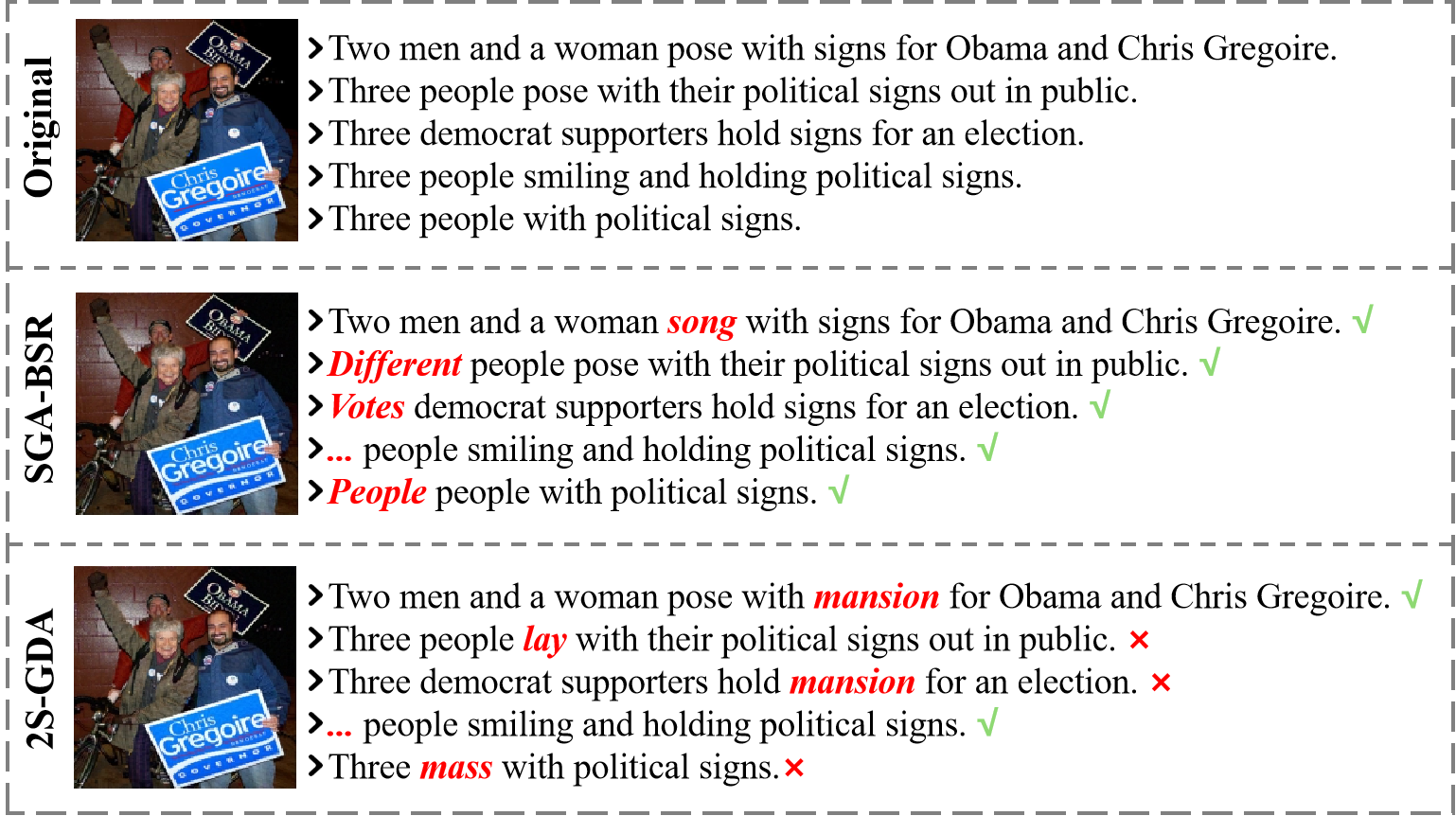}
        \caption{Visualization of adversarial examples on the image-text retrieval task (ALBEF→{CLIP}\textsubscript{ViT}).
        }
    \label{fig:3}
    \vspace{-2.5mm}
\end{figure}

\begin{figure}[t]
\centering
\scriptsize
\includegraphics[width=0.95\linewidth]{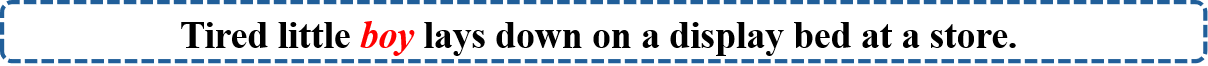}
\vspace{1pt}
\begin{tabular}{cccccc}
\includegraphics[width=0.14\linewidth]{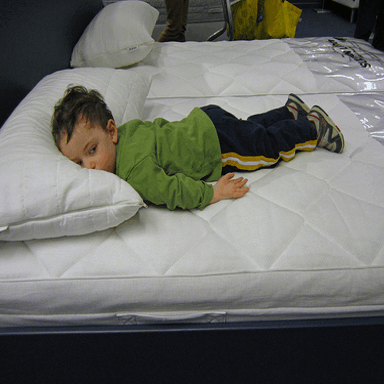} & 
\includegraphics[width=0.14\linewidth]{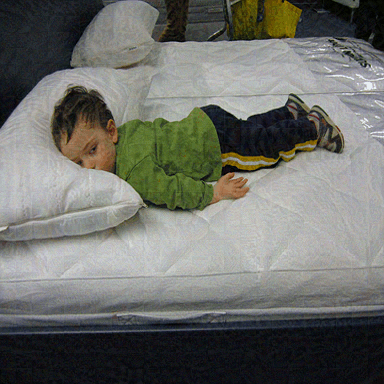} & 
\includegraphics[width=0.14\linewidth]{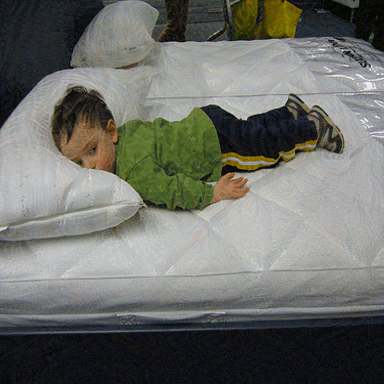} & 
\includegraphics[width=0.14\linewidth]{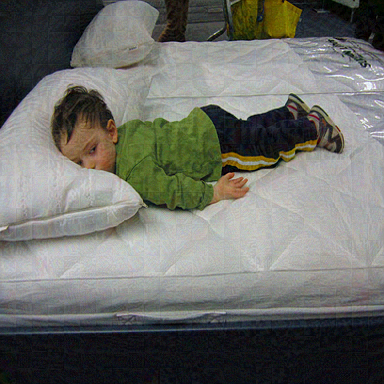} & 
\includegraphics[width=0.14\linewidth]{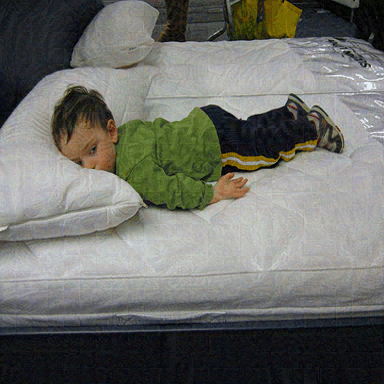} & 
\includegraphics[width=0.14\linewidth]{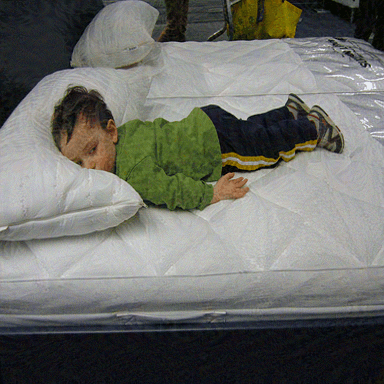} \\
\includegraphics[width=0.14\linewidth]{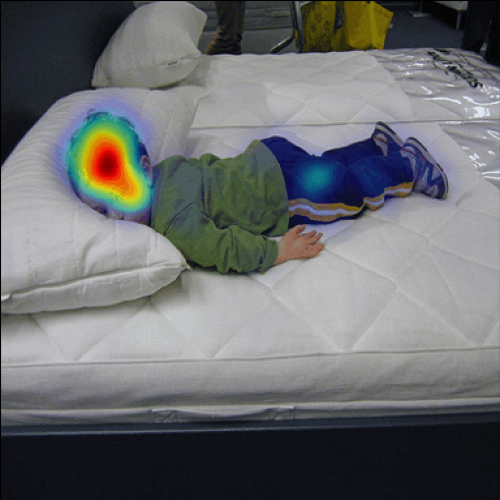} & 
\includegraphics[width=0.14\linewidth]{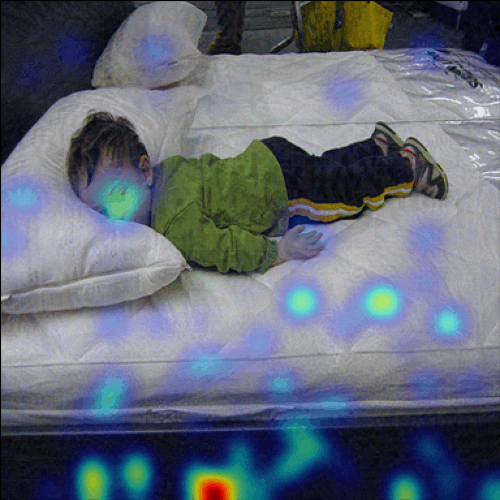} & 
\includegraphics[width=0.14\linewidth]{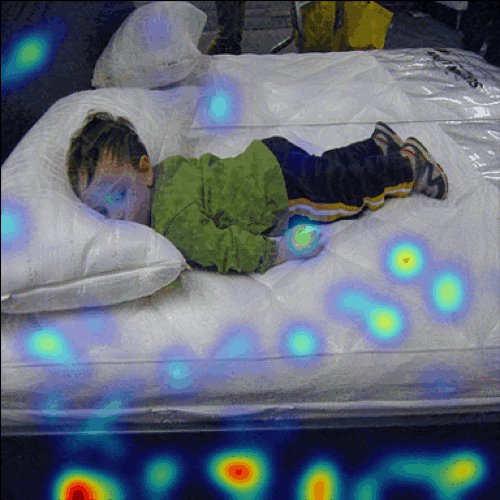} & 
\includegraphics[width=0.14\linewidth]{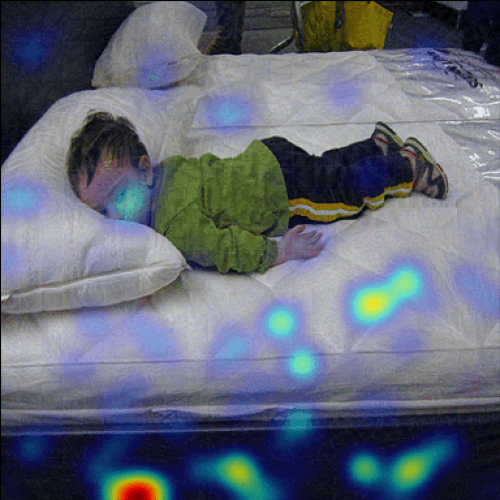} & 
\includegraphics[width=0.14\linewidth]{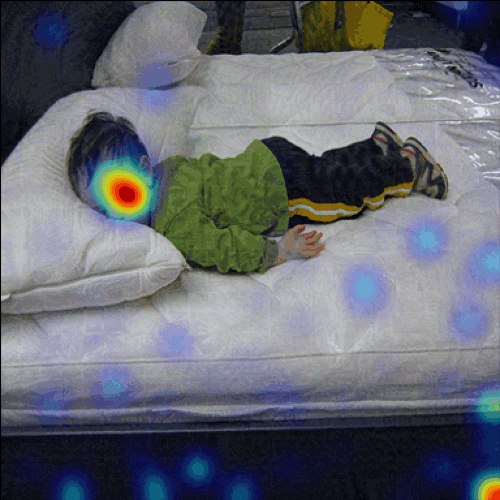} & 
\includegraphics[width=0.14\linewidth]{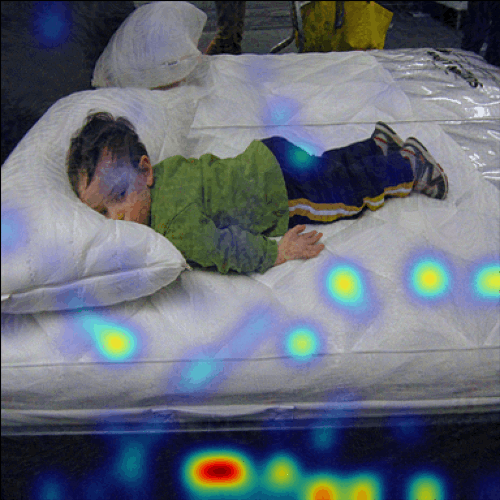}\\
\footnotesize{Raw} & \footnotesize{SGA} & \footnotesize{SGA-BSR} & \footnotesize{DRA} & \footnotesize{SA-AET} & \footnotesize{2S-GDA}
\end{tabular}
\vspace{-1mm}
\caption{Visualization of clean vs. adversarial examples. Rows show captions, images, and salient regions for “boy”. 
}
\label{fig:4}
\vspace{-2mm}
\end{figure}

In addition, several visualization results are provided. Fig.~\ref{fig:3} presents a visual comparison between SGA-BSR and our method on the image-text retrieval task. 
The notation “Model A → Model B” denotes that adversarial examples are generated by “Model A” and then evaluated on “Model B”.
The red text indicates the modified words, while the red crosses and green checks denote incorrect and correct retrieval matches, corresponding to successful and failed attacks, respectively.
The results demonstrate that 2S-GDA more effectively disrupts visual-text alignment. Fig.~\ref{fig:4} highlights the salient regions for the word “boy” using ALBEF on clean and adversarial images. Compared to baselines, 2S-GDA shifts attention away from the correct region, demonstrating stronger interference with semantic alignment.



\setlength{\tabcolsep}{2.2pt}
\begin{table}[t]
    \centering
  
    \caption{
        ASR (\%) of our method combined with baseline methods. The
        adversarial examples are generated by ALBEF.}
        
    \fontsize{5.5pt}{7.5pt}\selectfont 
    \begin{tabular}{ccccccccc}
        
        \toprule
      
        \multirow{2}{*}{Attack} & \multicolumn{2}{c}{{ALBEF}} & \multicolumn{2}{c}{{TCL}} & \multicolumn{2}{c}{{CLIP}\textsubscript{ViT}} & \multicolumn{2}{c}{{CLIP}\textsubscript{CNN}} \\
        \cmidrule(lr){2-3} \cmidrule(lr){4-5} \cmidrule(lr){6-7} \cmidrule(lr){8-9}
    
        & TR R@1 & IR R@1 & TR R@1 & IR R@1 & TR R@1 & IR R@1 & TR R@1 & IR R@1 \\
        \midrule
        DRA-BSR  & \cellcolor{gray!10}99.79 & \cellcolor{gray!10}99.93 & 91.04 & 91.33 & 46.75 & 57.25 & 50.45 & 59.59 \\
        2S-GDA-DRA (ours) & \cellcolor{gray!10}\textbf{100.00} & \cellcolor{gray!10}\textbf{100.00} & \textbf{93.78} & \textbf{94.07} & \textbf{52.64} & \textbf{64.21} & \textbf{56.58} & \textbf{67.10} \\
        \midrule
        SA-AET-BSR & \cellcolor{gray!10}99.79 & \cellcolor{gray!10}\textbf{99.95} & 96.21 & 96.21 & 75.21 & 79.54 & 77.14 & 80.27 \\
        2S-GDA-SA-AET (ours) & \cellcolor{gray!10}\textbf{100.00} & \cellcolor{gray!10}99.93 & \textbf{98.00} & \textbf{97.88} & \textbf{80.98} & \textbf{84.89} & \textbf{81.61} & \textbf{84.36} \\
        \bottomrule
    \end{tabular}

    \label{tab:2}
\end{table}


\begin{table}[!]
    \centering
     \caption{Ablation results of different modules on image-text retrieval across three attack frameworks.}
     \scriptsize 
        \begin{tabular}{lcccccc}
            \toprule
            & \multicolumn{2}{c}{{SGA-BSR}} & \multicolumn{2}{c}{{DRA-BSR}} & \multicolumn{2}{c}{{SA-AET-BSR}} \\
            \cmidrule(lr){2-3} \cmidrule(lr){4-5} \cmidrule(lr){6-7}
            & TR R@1 & IR R@1 & TR R@1 & IR R@1 & TR R@1 & IR R@1 \\
            \midrule
            w/o 2S    & 64.91 & 70.94 & \textbf{53.50}  & 62.50  & 80.25 & 84.66 \\
            w/o CTE   & 71.90 & 75.55 & 49.08 & 59.79 & 77.67 & 81.12 \\
            w/o GAR   & 71.66 & 76.90 & 52.39 & 62.18 & 79.63 & 81.86 \\
            Completed &\cellcolor{gray!10} \textbf{74.72} &\cellcolor{gray!10} \textbf{79.64} &\cellcolor{gray!10} 52.64 &\cellcolor{gray!10} \textbf{64.21} &\cellcolor{gray!10} \textbf{80.98} &\cellcolor{gray!10} \textbf{84.89} \\
            \bottomrule
            \label{tab:3}
        \end{tabular}%
        \vspace{-1mm}
\end{table}

\begin{figure}[!]
    \centering
    \includegraphics[width=1\linewidth]{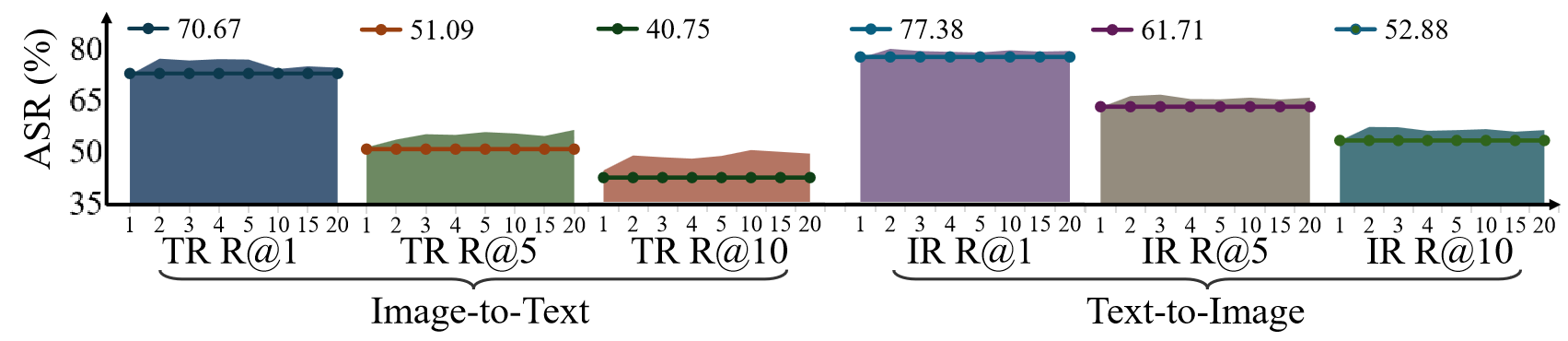}
    \caption{Comparison of attack success rates under different number of influential tokens $k$ in GAR, with the baseline (w/o GAR) included for reference (ALBEF→CLIP\textsubscript{ViT}).}
    \label{fig:5}
    \vspace{-2.5mm}
\end{figure}

\vspace{-1.5mm}
\subsection{Ablation Studies}
\vspace{-0.5mm}

Firstly, we perform ablation studies to assess the impact of the proposed components: the two-stage pipeline (2S), Candidate Text Expansion (CTE), and Globally-Aware Replacement (GAR). As shown in Table~\ref{tab:3}, removing any module leads to a noticeable performance drop, confirming their effectiveness and complementarity. Among them, excluding 2S causes the most significant degradation, highlighting the importance of avoiding repeated textual perturbation. 

Secondly, we study the effect of the influential tokens $k$ in GAR. Fig.~\ref{fig:5} presents the results on ALBEF→CLIP\textsubscript{ViT}, while the results on ALBEF→TCL and ALBEF→CLIP\textsubscript{CNN} are provided in Fig. A4 of the supplementary material. As observed, the ASR peaks around $k=3$ and then declines, which may be attributed to excessive semantic distortion.

\section{Conclusion}
In this work, we propose 2S-GDA, a two-stage attack framework by introducing the globally-diverse strategy. Extensive experiments demonstrate that 2S-GDA consistently outperforms state-of-the-art baselines, particularly in black-box settings. Moreover, our proposed framework exhibits better extensibility and can be easily combined with existing methods to further improve transferability.




%
\begin{spacing}{0.95}
\bibliographystyle{IEEEbib}
\bibliography{strings,refs}
\end{spacing}

\end{document}